\documentclass[conference]{IEEEtran}
\IEEEoverridecommandlockouts
\usepackage{cite}
\usepackage{amsmath,amssymb,amsfonts}
\usepackage{algorithmic}
\usepackage{graphicx}
\usepackage{textcomp}
\usepackage{xcolor}
\usepackage{url}
\def\BibTeX{{\rm B\kern-.05em{\sc i\kern-.025em b}\kern-.08em
		T\kern-.1667em\lower.7ex\hbox{E}\kern-.125emX}}
\begin{document}
	
\title{Regression Oracles and Exploration Strategies for \\ Short-Horizon Multi-Armed Bandits
\thanks{Work partially funded by the NSF (Grant Number: IIS-1816470)}
}

\author{\IEEEauthorblockN{Robert C. Gray}
\IEEEauthorblockA{\textit{College of Media Arts \& Design} \\
\textit{Drexel University}\\
Philadelphia, PA, USA \\
robert.c.gray@drexel.edu}
\and
\IEEEauthorblockN{Jichen Zhu}
\IEEEauthorblockA{\textit{College of Media Arts \& Design} \\
\textit{Drexel University}\\
Philadelphia, PA, USA \\
jichen.zhu@gmail.com}
\and
\IEEEauthorblockN{Santiago Onta\~n\'on*\thanks{*Currently at Google}}
\IEEEauthorblockA{\textit{College of Computing \& Informatics} \\
\textit{Drexel University}\\
Philadelphia, PA, USA \\
so367@drexel.edu}
}


\maketitle

\begin{abstract}
This paper explores multi-armed bandit (MAB) strategies in very short horizon scenarios, i.e., when the bandit strategy is only allowed very few interactions with the environment. This is an understudied setting in the MAB literature with many applications in the context of games, such as player modeling. Specifically, we pursue three different ideas. First, we explore the use of {\em regression oracles}, which replace the simple average used in strategies such as $\epsilon$-greedy with linear regression models. Second, we examine different exploration patterns such as {\em forced exploration} phases. Finally, we introduce a new variant of the UCB1 strategy called UCBT that has interesting properties and no tunable parameters. We present experimental results in a domain motivated by exergames, where the goal is to maximize a player's daily steps. Our results show that the combination of $\epsilon$-greedy or $\epsilon$-decreasing with regression oracles outperforms all other tested strategies in the short horizon setting.
	
\end{abstract}

\begin{IEEEkeywords}
multi-armed bandit, player modeling, machine learning, linear regression, reinforcement learning
\end{IEEEkeywords}

\section{Introduction}\label{sec:introduction}

Multi-armed bandits (MABs) refer to a class of sequential decision problems~\cite{thompson1933likelihood, robbins1952some} where an agent attempts to maximize rewards received by repeatedly choosing one out of a set of actions, each with unknown and stochastic rewards. MAB strategies aim to balance the need for exploration against the benefits of exploiting the actions believed to be the most rewarding. Specifically, this paper focuses on the problem of defining MAB strategies that target scenarios with a very short {\em horizon}, i.e., those in which the agent gets to choose an action a very small number of times.

Short horizon bandits arise in many real-world situations. For example, consider the problem of player modeling~\cite{yannakakis2013, drachen2009player,valls2015exploring}, where software systems aim to model or classify players to provide them with individualized experiences. In our previous work~\cite{gray2020player}, we demonstrated that MAB strategies can be useful in automatically determining the best options to present a user, based on that user's traits, as the MAB strategy observes the user's reactions to the different options. However, if MAB strategies are to be useful for player modeling, they need to adapt to users quickly within a small number of iterations. This is in contrast to standard analyses of MAB strategies, which are typically studied in the limit, where the number of iterations is very large. 

To address this gap, in this paper we study MAB strategies that attempt to converge very quickly. To do so, we study three different ideas: (1) integrating linear regression models into the {\em oracle} of the MAB to more accurately predict future rewards given past information; (2) forced exploration patterns 
that focus on pure exploration initially before engaging in the bandit strategy;
and finally (3) we compare different MAB strategy families, including a new variant of the UCB strategy~\cite{Auer2002} we call {\em UCBT} that can automatically adjust its exploration constant based on observations rather than requiring it to be externally tuned. Our empirical results show that all three ideas can help significantly in the short horizon scenario and that although our best performing strategies are variants of $\epsilon$-greedy and $\epsilon$-decreasing incorporating ideas (1) and (2), UCBT compares favorably to UCB when the exploration constant is not properly tuned.

The remainder of this paper is structured as follows.  First, we present some related research on MAB strategies that face similar challenges and constraints. After that, we introduce the motivating scenario and simulator we use to compare our approaches. Then, we present our approach toward addressing the short-horizon MAB problem. Finally, we examine the results of MAB performance in a simulator designed to mimic human walking behavior patterns to further observe the potential for regression-based MABs. 

\section{Background and Related Work}\label{sec:related-work}

A multiarmed bandit (MAB) problem~\cite{lai1985asymptotically} is a sequential decision problem in which an agent needs to select one of $k$ actions (called {\em arms}) sequentially over the course of $h$ time steps (or {\em horizon}). At each time step, the agent receives a reward based on the arm chosen. Neither the reward distributions nor the expected values of the arm rewards $\rho$ are known beforehand. The goal of the agent is to maximize the obtained (cumulative) reward. Thus, at each iteration, the agent must choose between {\em exploiting} (selecting the arms that have so far performed the best) or {\em exploring} (selecting other arms to gather additional data about them). 

In the most common instantiation of the MAB problem, known as the {\em stochastic bandit problem}, each of the arms available to the agent holds a static, unknown, underlying reward distribution from which a reward is randomly drawn~\cite{kuleshov2000}. MAB strategies usually assess the expected reward of each arm based on past iterations via some estimation process (which we will call an {\em oracle}). In the case of stochastic bandits, typical strategies estimate descriptive statistics of observed rewards such as mean $\mu$ and standard deviation $\sigma$. These values provide the basis for predictions in popular MAB approaches such as the $\epsilon$-based (e.g., {\em $\epsilon$-greedy}, {\em $\epsilon$-decreasing}) and {\em Upper Confidence Bound} (UCB) families of strategies~\cite{lattimore2018bandit}. Two lines of work on MAB strategies are relevant to the work we present in this paper: (1) MAB strategies for domains with contextual or non-stationary rewards, and (2) MAB strategies for domains with very short horizons, briefly described below.

In this standard stochastic bandit formulation, rewards are assumed to depend exclusively on the selected arm. In other words, the problem is {\em stateless} and arm choices in previous iterations do not affect rewards obtained in the current iteration. However, as elaborated in Section~\ref{sec:simulation-environments}, the domain we use in this paper does not strictly satisfy this property; we focus on social exergames, where the goal is to maximize the number of steps of the players. This metric appears to correlate with several other factors, such as the day of the week, activity levels of previous days, and others. In particular, three variations of the MAB problem that are particularly relevant to model this are {\em contextual bandits}~\cite{langford2007epoch}, {\em non-stationary bandits}~\cite{kocsis2006discounted}, and {\em restless bandits}~\cite{whittle1988restless}. In contextual bandits, the stochastic reward function is assumed to depend on an external {\em context vector}, which is observed by the agent before having to choose an arm. Non-stationary and restless bandits model the scenario where the stochastic reward functions of all the arms change over time. For example, {\em Discounted UCB}~\cite{kocsis2006discounted} adds a discount factor for observations far in the past, and Besbes et al.~\cite{besbes2019optimal} propose the use of a ``dynamic oracle'' that observes the way rewards change over the horizon and considers that variation when selecting arms.

Concerning domains with very short horizons, although some isolated pieces of work exist, the problem has not received much attention in the literature. A recent approach by Tomkins et al.~\cite{tomkins2020rapidly} considers very short horizons when using bandits to select personalized policies for users. To address the problem, they exploit knowledge of rewards from trials with other users.  This ``intelligent pooling'' strategy assesses how similar or dissimilar two users are to determine how much one user's observed rewards should be applied to the others. Although this is a promising approach, we do not employ such technique in this paper because we do not assume access to data from other users.


\section{Simulation Environments}\label{sec:simulation-environments}

The main challenge addressed in this paper (short-horizon MABs) is motivated by deploying MABs in domains where an MAB strategy needs to select with low frequency (e.g., once a day) a scenario or intervention for a human player, hoping to have a specific desired effect on the player. For example, consider exergames~\cite{zhu2018towards, gray2018}, where the goal is to motivate players to do more exercise. In these domains, we cannot expect a human to play the game for thousands or even hundreds of days. Therefore, if the MAB strategy is to be effective, it must learn to personalize the experience for the player within a very short time. 

Motivated by this problem and focusing on a domain where we want to maximize the number of steps walked by a player, we designed two simulation environments to evaluate the effectiveness of our approaches. Because we are using simulators to evaluate the different MAB strategies of our study, it is essential that the simulators reflect some of the important features that would be encountered when dealing with the real-world domain. The remainder of this section describes the design and rationale for these simulators.

\subsection{Human Step Behavior Modeling}

In both simulation environments, the task of the agent is to maximize the daily steps of the {\em virtual players} in an exergame or health study. The only thing that can be manipulated by the agent is the selection among three variants of the {\em intervention} that the player will see each day. The task is formulated as an MAB problem in which the agent will sequentially select from $k=3$ arms $A_{1,...,3}$ one choice for each player to maximize the reward $\rho_t$ (i.e., daily steps) observed each day $t$ over the course of the study's {\em horizon} $h$.

\begin{figure}[t!]
	\includegraphics[width=\columnwidth]{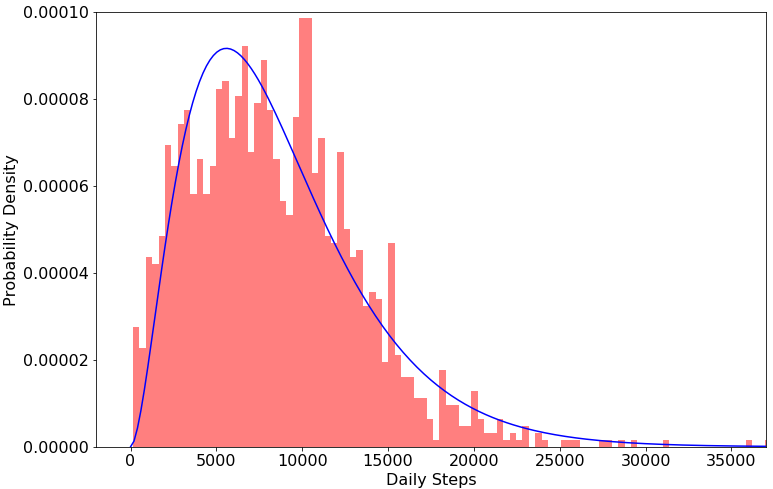}
	\centering
	\caption{Histogram of daily step recordings collected from participants in a Mechanical Turk experiment~\cite{furberg2016} ($n=1665$) overlayed with the probability density function for $\Gamma(k=2.8,\theta=3100)$.}
	\label{fig:mech-turk-original}
\end{figure}

To generate a realistic reward from the virtual players in our simulators, we used data from a Mechanical Turk experiment conducted in 2016 by Furberg et al.~\cite{furberg2016} that collected participants' daily steps measured by a Fitbit. After removing outliers (values less than 100, $n=20$), we graph the data ($n=1665$) as a histogram in Figure~\ref{fig:mech-turk-original}. After confirming the data to be not normally distributed, we fit a Gamma distribution based on literature regarding human walking patterns and other intermittent behaviors~\cite{orendurff2008, guo2011}. The probability density function for $\Gamma(k=2.8,\theta=3100)$ is included in Figure~\ref{fig:mech-turk-original}.

\subsection{Stationary Step Simulator}\label{sec:simulator1}

In our first simulator, which we call the {\em stationary step simulator}, we generate daily steps for each virtual player by drawing randomly from this Gamma distribution. However, we note that the nature of human walking behavior is not fully replicated by this approach, even if we are able to generate data that matches this distribution in aggregate. Therefore, we pursue a higher degree of fidelity in our second simulator, discussed below.

\subsection{Pattern Step Simulator}\label{sec:simulator2}

Many human activities are habitual, bound to routine, and exhibit cyclical patterns~\cite{kielhofner2002model}. For example, if presuming a societal convention of planning routine activities within a 7-day week, it's reasonable to expect that an individual's walking patterns on the first day of that week would correlate with their walking patterns on the first day of other weeks. Thus, in our second simulator, which we call the {\em pattern step simulator}, we aim to exhibit this correlative and cyclical nature of daily step behavior while maintaining adherence to the overall expected distribution.

Specifically, to estimate the degree to which prior days' steps might predict the current day's steps, we constructed a regression model from the sequential step data using the seven days prior to a given step count as the features of the regression:
\begin{equation}\label{eq:s-t-calculation}
S_t = C + \sum_{i=1}^{7}\beta_iS_{t-i}
\end{equation}

An Ordinary Least Squares (OLS) regression revealed the coefficients $\beta$ for these features, with which we employed backward elimination to construct a minimal model, sequentially removing the features that showed the least statistical significance until only those with $p<0.05$ remained.  All features survived this process with $p<0.001$ except $S_{t-5}$, the number of steps five days prior to the current day.  The resulting coefficients are presented in Table~\ref{tab:ols-regression-weights}. 

To prime the steps for the first seven days, we simply pull from a Gamma distribution with $k=2.8$ and $\theta=3100$ as with the stationary step simulator. After that (i.e., when $t > 7$), the step count is generated as follows:
\begin{equation}\label{eq:pattern-s-t-calculation}
S_t = C + \sum_{i=1}^{7}\beta_iS_{t-i} + g,\quad g\sim\Gamma(k, \theta)
\end{equation}

\noindent where $\beta$ is the set of coefficients listed in the Mechanical Turk column of Table~\ref{tab:ols-regression-weights}, $\beta_5=0$, $C=-3000$, $k=1.1$, and $\theta=3500$.
If $S_t$ is negative, it is discarded and a new value is sampled.

\begin{figure}[t!]
	\includegraphics[width=\columnwidth]{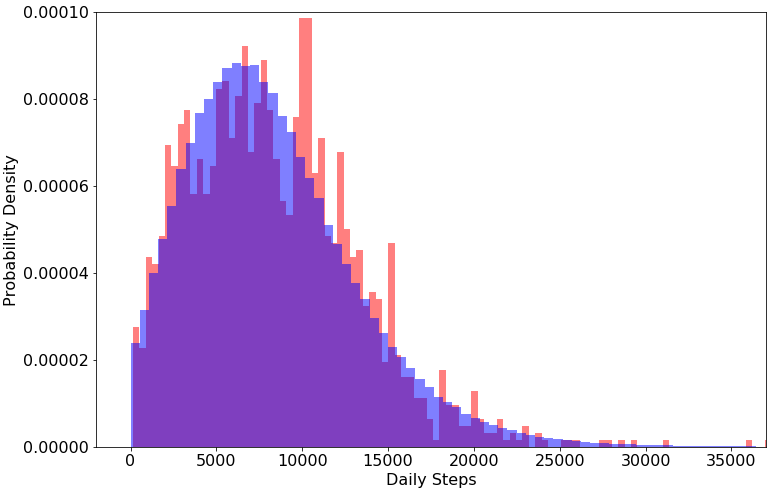}
	\centering
	\caption{Histogram (red) of daily step recordings collected from participants in a Mechanical Turk experiment~\cite{furberg2016} ($n=1665$) overlayed with a histogram (blue) of the steps generated by the pattern step simulator ($n=500k$).}
	\label{fig:mech-turk-gamma}
\end{figure}

To verify the accuracy of the simulator, results of an experiment of 500k daily steps is presented in Figure~\ref{fig:mech-turk-gamma}, where a histogram of the resulting probability density overlays a histogram of the original Mechanical Turk data. Additionally, we performed the same regression procedure with backward elimination on this data to verify that it resulted in a model with precisely the same significant features and similar coefficients. The results can be seen in the third column of Table~\ref{tab:ols-regression-weights}, verifying that the trends in our generated data match those observed in the real-world data.

We thus have a step simulator that both 1) reflects the observed real human step distribution and 2) maintains relative correlations observed in the real human daily step data. Let us now formulate the MAB problem used for evaluating our approach.

\begin{table}[t!]
	\caption{Significant linear regression coefficients ($p < 0.001$) for step data resulting from Mechanical Turk~\cite{furberg2016} ($n=1665$) and Step Simulator ($n=500k$) experiments. $S_{t-5}$ intentionally omitted.
	}
	\begin{center}
		\begin{tabular}{|c|c|c|}
			\hline
			&\multicolumn{2}{|c|}{\textbf{OLS Regression Coefficients $\beta$}} \\
			\cline{2-3} 
			\textbf{Feature} & {\bf {\em Mechanical Turk} }&{\bf {\em Step Simulator}} \\
			\hline
			$S_{t-1}$ & 0.2599 & 0.2540 \\ 
			$S_{t-2}$ & 0.0984 & 0.0952 \\ 
			$S_{t-3}$ & 0.0851 & 0.0827 \\ 
			$S_{t-4}$ & 0.1337 & 0.1274 \\ 
			$S_{t-6}$ & 0.1300 & 0.1281 \\ 
			$S_{t-7}$ & 0.1833 & 0.1826 \\ 
			\hline
		\end{tabular}
		\label{tab:ols-regression-weights}
	\end{center}
\end{table}

\subsection{MAB Arms}

The step value $S_t$ generated for a simulated user on a given day $t$ is influenced by the arm selected by the MAB strategy. Specifically, each arm is defined by an {\em adjustment range} $[l,h]$. When the MAB strategy selects an arm, an {\em adjustment value} $r$ is uniformly sampled from the interval $[l,h]$, and the observed reward $\rho_t$ is calculated as follows:
\begin{equation}\label{eq:resulting-steps}
\rho_t =  S_t * (1 + r)
\end{equation}

The intuition is that these adjustment values would represent the degree to which a selected intervention could influence the number of steps the player would walk. Specifically, positive adjustment values indicate that the player would walk more steps as a consequence of the intervention, and negative values indicate the opposite. In this scenario, we devised a bandit that would select from three potential interventions, and the adjustment ranges for each of these three arms (A, B, and C) are shown on the third column of Table~\ref{tab:mab-arms}.

\begin{table}[t!]
	\caption{ MAB Arms }
	\begin{center}
		\begin{tabular}{|c|c|c|}
			\hline
			{\bf Arm Name} & {\bf Oracle Value $O_a$} & {\bf Adjustment Range} \\
			\hline
			A & -.2 & [-0.2, 0.0] \\
			B & -.1 & [-0.1, 0.1] \\
			C & 0 & [0.0, 0.2]  \\
			\hline
		\end{tabular}
		\label{tab:mab-arms}
	\end{center}
\end{table}


\section{Approach}\label{sec:approach}

To address the short horizon problem that arises in our motivating domain, we explored three ideas: (1) improving the oracle used by the MAB strategies, (2) employing fixed exploration patterns, and (3) exploring the efficacy of a collection of MAB strategy families, including a new adaptive variant of UCB which we call {\em UCBT}. We describe these three ideas below.

\subsection{Regression Oracle}\label{sec:approach-regression-oracle}

A typical MAB strategy, such as $\epsilon$-greedy, will remember the rewards received for each pull of each arm and estimate the expected reward of a given arm by calculating the mean of those previously observed rewards. We call this calculation procedure the {\em oracle} used by the strategy to predict the expected reward of a given arm at the current time step. The policy for such a strategy only needs to compare the expected values for the rewards of all arms to determine which it should pull to receive the greatest reward.  Other strategies like UCB1 additionally incorporate an estimation of the confidence interval to choose which arm to pull at each time step.


However, we note that these types of oracles make two assumptions inherent to the standard MAB problem formulation: (1) they assume reward distributions are stationary, and (2) they assume reward distributions of different arms are unrelated. Both of these assumptions are violated in our motivating domain (and in any domain involving repeatedly interacting with the same human subject). Standard approaches for non-stationary rewards (see Section~\ref{sec:related-work}) such as weighted averages with decaying weight for past observations would not be useful in our short horizon setting; they would not be sample-efficient enough, nor would they leverage the type of patterns observed in human walking behavior.

To address this, we propose replacing the standard oracle used in MAB strategies with a {\em regression oracle} that makes a prediction based on not only the past observed rewards for a given arm, but on all past observed rewards.
We implement this regression oracle by collating the previously observed rewards into a linear regression model, one that is capable of potentially capturing some of the temporal patterns we anticipate to be present in the generated step data. We expect this model to provide more accurate predictions of the expected values of arm rewards than simple means of past observations.

Specifically, at a given time step $t$ our regression model takes an input vector $\hat{x}_{t,a} = (\rho_{t-1},... , \rho_{t-m}, O_a)$, containing the observed rewards for the past $m$ iterations ($m = 7$ in our experiments), and $O_a$, which indicates the arm $a$ for which we desire to generate a prediction (values of $O_{a}$ for each of the 3 arms in our experiments are shown in Table~\ref{tab:mab-parameter-values}). The model is trained to predict $\rho_{t,a}$, the expected reward we would obtain when pulling arm $a$ at time $t$.


Each time an arm is pulled and a reward observed, an OLS regression is performed on all the past observations to determine an updated set of regression coefficients $\beta$.
The next time ($t$) an arm must be selected, a predicted reward value $\hat{\rho}_{t,a}$ is calculated for each arm $a$ as follows:
\begin{equation}\label{eq:regression-prediction}
\hat{\rho}_{t,a} =\sum_{i=1}^{m}\beta_i\hat{x}_{t,a,i}
\end{equation}
%
Notice that this is a strict generalization over calculating the mean; if there are no temporal correlations that can be exploited, the $\beta$ parameters corresponding to the reward for the past $m$ time steps will converge to either: (a) 0, and the coefficient for the $O_a$ parameter will converge to the mean reward for arm $a$ divided by $O_a$, or (b) a uniform distribution, and the coefficient of the $O_a$ term will be used to distinguish the reward of each arm. However, if temporal patterns or correlation among arms do exist, this oracle will be able to exploit them. More elaborate regression models could be devised, but given that we are focusing on the short horizon scenario and we cannot expect the model to be trained with more than a few dozen data points, simple linear regression is justified.

Also, we note that this formalization is very reminiscent of a contextual bandit, where we could consider the rewards observed over the past $m$ steps as the context vector. Moreover, we are also aware that because the observed rewards in the previous time steps depend on the arm being pulled, we are violating one of the basic assumptions of the MAB problem formulation--namely, that the bandit should be stateless, and previous arm selections should not affect future rewards. Our problem therefore resembles more a reinforcement learning setting than a bandit setting. However, we choose to approach it with MAB strategies due to the focus on short horizon; introducing the notion of state would imply more parameters to estimate in the model and would thereby decrease sample efficiency.


\subsection{Exploration Patterns}\label{sec:approach-exploration-patterns}

Most bandit strategies, such as $\epsilon$-greedy, rely on stochastic exploration of the arms. However, in short horizon settings, stochastic exploration might not be appropriate, and alternative strategies that perform heavier exploration at the beginning have been proposed. In order to determine the behavior of different exploration patterns, we compared a range of MAB strategies in both step simulators:
\newcounter{mablist}
\begin{enumerate}
	\item UCB~\cite{Auer2002}: selects the arm with the highest potential reward based on confidence intervals around the average of past rewards.
	\item $\epsilon$-greedy: selects the ``best'' performing arm except when (with probability $\epsilon$) it {\em explores} by choosing a random arm.
	\item $\epsilon$-decreasing: similar to $\epsilon$-greedy, except with exploration probability $1/t^\epsilon$.
	\setcounter{mablist}{\value{enumi}}
\end{enumerate}

Additionally, for each of the strategies we explored the idea of {\em forced exploration} periods in which the MAB strategy is not permitted to engage its policy until each arm has been explored a specified number of times. During the forced exploration period, all arms are pulled an equal number of times, but the order in which they are pulled remains random. It is worth noting that we did not include the $\epsilon$-first strategy, as we view $\epsilon$-first to be simply $\epsilon$-greedy with forced exploration, a modification that could be similarly applied to any MAB strategy. Moreover, although this is only strictly necessary for UCB strategies, in our experiments all strategies (UCB, $\epsilon$-greedy, and $\epsilon$-decreasing) still undergo a forced exploration period of one pull per arm (in random order) to establish an estimate for the mean reward for each arm. Thus, in our experiments, when we label a strategy with {\em forced exploration}, we refer to a forced exploration period of four pulls per arm before engaging the strategy's policy, whereas {\em no forced exploration} refers to forced exploration of just one initial pull per arm.

\begin{table}[t!]
	\caption{Parameters for strategies in both simulation environments.}
	\begin{center}
		\begin{tabular}{|c|c|c|}
			\hline
			{\bf Strategy} & {\bf Stationary Step Simulator} & {\bf Pattern Step Simulator} \\
			\hline
			UCB1                  & $C=2500$        & $C=1600$ \\
			$\epsilon$-greedy     & $\epsilon=0.11$ & $\epsilon=0.03$ \\
			$\epsilon$-decreasing & $\epsilon=0.7$  & $\epsilon=1.0$ \\
			\hline
		\end{tabular}
		\label{tab:mab-parameter-values}
	\end{center}
\end{table}

\subsection{UCBT Strategy}\label{sec:approach-ucbt}

The UCB1 algorithm leverages the Chernoff-Hoeffding bounds~\cite{Auer2002} to provide an estimate of the upper confidence bound for a reward distribution derived from the observed rewards and a variance factor based on the number of observations.  For reward distributions not confined to $[0, 1]$, often an additional {\em exploration factor} $C$ is also included to empower the variance factor to influence the oracle's valuation to a degree appropriate for the scale of rewards:
\begin{equation}\label{eq:ucb1-calculation}
UCB1_a = \bar{x}_a + C\sqrt{\frac{2\ln{t}}{{n_a}}}
\end{equation}

\noindent where $\bar{x}_a$ is the mean of the rewards observed so far for arm $a$, $n_a$ is the number of times arm $a$ has so far been selected, and $t$ is the current time step.

UCB1 is a popular strategy but may hold disadvantages in some applications. For instance, not every scenario can normalize rewards to $[0, 1]$, such as when rewards have no upper limit. In these cases, the $C$ factor must be specially tuned to the scenario in order for the strategy to work effectively.  However, some applications do not provide a means for pre-sampling and tuning this factor, in which case UCB1 may underperform.

In our scenario, rewards are unbounded, and thus to address this problem, we designed an alternative strategy {\em UCBT} that considers traditional statistical confidence bounds using sample variance. Because the variance factor calculation includes this consideration for the scale of rewards, the additional exploration factor $C$ is no longer required. UCBT does presume normality in the underlying reward distribution and therefore constructs a confidence interval using Student's $T$ distribution. The critical value $t^*$ used in our experiments was drawn from a $T$-distribution lookup table with a 1-sided critical region and 99\% confidence. We used a factor from a standard normal distribution ($t^*=2.326$) when degrees of freedom exceeded 200.

Specifically, on pull $t$, UCBT selects the arm $a$ for which the following yields the greatest value:
\begin{equation}\label{eq:ucbt-calculation}
UCBT_a = \bar{x}_a + t^* \frac{s_a}{\sqrt{n_a}}
\end{equation}

\noindent where $s_a$ is the sample standard deviation of the rewards so far observed for arm $a$. We must note that UCBT will not function properly when all observations are identical (i.e., sample variance is zero). Also, prior to engaging its policy, UCBT requires each arm to be selected two times in order to establish an estimate for confidence interval for each arm; this results in what is effectively a forced exploration period of two pulls per arm. 

The next section presents experimental results to evaluate each of the three ideas considered in this paper to address the short horizon bandit problem in our target domain.

\section{Experimental Evaluation}\label{sec:results}

To evaluate our ideas, we compared six different bandit strategies: UCB1, UCBT, $\epsilon$-greedy, $\epsilon$-decreasing, $\epsilon$-greedy with regression oracle ($\epsilon$-greedy regression) and $\epsilon$-decreasing with regression oracle ($\epsilon$-decreasing regression). For each of the strategies above that required tuning a parameter, we ran a series of experiments (each running 10 million times with a horizon $h=70$) over potential values for their respective parameters (i.e., $C$ or $\epsilon$) to find the value at which the strategy performed the best against our simulations. Table~\ref{tab:mab-parameter-values} contains the values we used in our experiments.



We conducted several experiments in which all six MAB strategies were evaluated, and we discuss the performance of the MAB strategies in the following contexts: (1) in the stationary step simulator, (2) in the pattern step simulator, (3) with a four-period forced exploration phase in both simulators, and (4) in an evaluation of UCB1 vs. UCBT.

All experiments were conducted with a horizon of $h = 70$ (i.e., 10 weeks if assuming daily interaction with a player), which is a significantly shorter horizon than what is usually considered in the bandit literature. For each strategy and experiment, we recorded the rewards observed at each of the 70 time steps of the experiment, and we report the average of 1 million runs.


\subsection{Results in the Stationary Step Simulator}\label{sec:results-distribution-simulation}

\begin{figure}[t!]
	\includegraphics[width=\columnwidth]{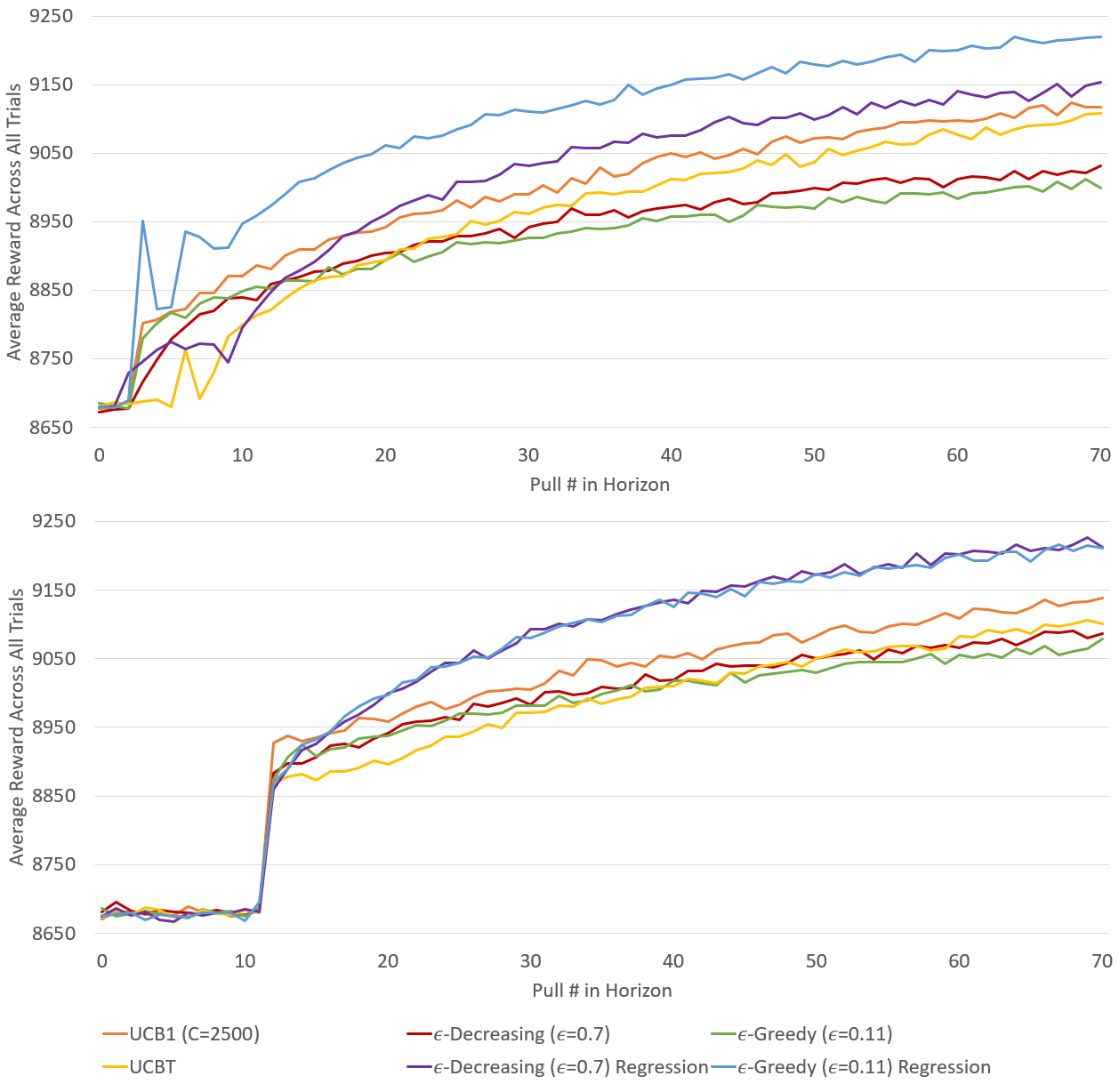}
	\centering
	\caption{Average reward over time for different MAB strategies without forced exploration (above) and with forced exploration (below) in the stationary step simulator.}
	\label{fig:gammaonly-combined}
\end{figure}

\begin{table}[t!]
	\caption{Average rewards in stationary step simulator ($h=70$)}
	\begin{center}
		\begin{tabular}{|c|c|c|c|c|}
			\hline
			& \multicolumn{2}{|c|}{\bf No Forced Exploration} & \multicolumn{2}{|c|}{\bf Forced Exploration} \\
			\cline{2-5}
			&   Overall & Last 7 Days & Overall & Last 7 Days \\ \hline
			UCB1                   & 8989.7       & 9114.5       & 8982.8       & 9129.6       \\
			UCBT                   & 8949.8       & 9096.0       & 8946.6       & 9097.8       \\
			$\epsilon$-greedy      & 8919.6       & 9001.9       & 8947.0       & 9064.1       \\
			$\epsilon$-decr.       & 8930.1       & 9022.0       & 8956.7       & 9083.2       \\
			$\epsilon$-greedy reg. & {\bf 9087.4} & {\bf 9216.4} & 9034.7       & 9207.8       \\
			$\epsilon$-decr. reg.  & 9003.7       & 9141.3       & {\bf 9036.7} & {\bf 9213.8} \\
			\hline
		\end{tabular}
		\label{tab:results-distribution}
	\end{center}
\end{table}

The reward distribution in the stationary step simulator is fixed, and the steps generated one day do not hold any correlations to steps from previous days. Average rewards across all steps from $t = 1$ to $t = 70$ for each MAB strategy (without forced exploration) can be found in the left columns of Table~\ref{tab:results-distribution}, and average reward over time is shown in the upper graph in Figure~\ref{fig:gammaonly-combined}.

\begin{table}[t!]
	\caption{Average rewards in Pattern Step Simulation ($h=70$)}
	\begin{center}
		\begin{tabular}{|c|c|c|c|c|}
			\hline
			& \multicolumn{2}{|c|}{\bf No Forced Exploration} & \multicolumn{2}{|c|}{\bf Forced Exploration} \\
			\cline{2-5}
			&   Overall & Last 7 Days & Overall & Last 7 Days \\ \hline
			UCB1                   & 8538.3       & 8586.0       & 8542.2       & 8616.6       \\
			UCBT                   & 8506.4       & 8597.2       & 8500.0       & 8594.6       \\
			$\epsilon$-greedy      & 8525.6       & 8560.4       & 8532.8       & 8602.7       \\
			$\epsilon$-decr.       & 8531.4       & 8577.4       & 8531.6       & 8612.0       \\
			$\epsilon$-greedy reg. & {\bf 8648.7} & {\bf 8713.6} & 8606.8       & 8712.0       \\
			$\epsilon$-decr. reg.  & 8607.6       & 8695.9       & {\bf 8609.4} & {\bf 8724.1} \\
			\hline
		\end{tabular}
		\label{tab:results-pattern}
	\end{center}
\end{table}

\begin{figure}[t!]
	\includegraphics[width=\columnwidth]{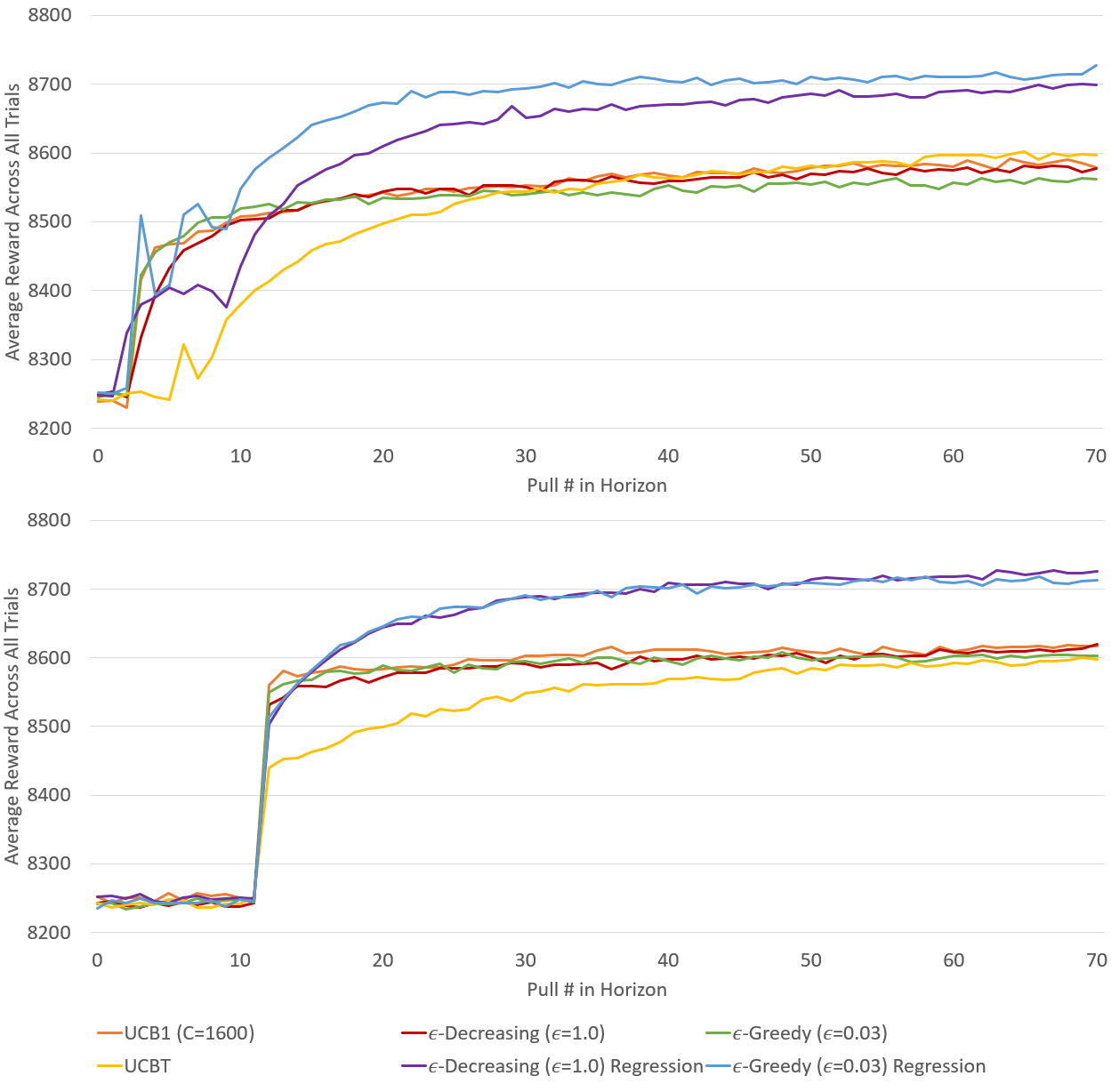}
	\centering
	\caption{Average reward over time for different MAB strategies without forced exploration (above) and with forced exploration (below) in the pattern step simulator.}
	\label{fig:realsteps-combined}
\end{figure}

The first thing we observe is that even in the stationary step simulator where there are no temporal correlations among steps, the strategies with the regression oracle still outperformed the other four strategies. In strategies that do not use the regression oracle, the expected rewards of each arm are calculated independently from each other. However, in the case of the regression oracle variants, rewards observed for {\em all} previous iterations (including those for other arms) are used for estimating expected reward. Since in our scenario the rewards of different arms are correlated (i.e., they all correspond to the number of daily steps for the same player), this enables the regression oracle to estimate the expected reward of each arm earlier and more accurately.




\subsection{Results in the Pattern Step Simulator}\label{sec:results-pattern-step-simulation}

As discussed above, the pattern step simulator attempts to structure a temporal correlation among daily steps that reflects the relationships observed in real human walking behavior. Average rewards across all steps without forced exploration in this simulator are shown in the left columns of Table~\ref{tab:results-pattern}, and average reward over time is shown in the upper graph of Figure~\ref{fig:realsteps-combined}.

In the experiment without forced exploration, the regression oracle strategies struggle in the early steps as the regression models begin to populate with observations. However, these two strategies very quickly overtake the other four strategies once enough data has been collected. Interestingly, notice that this happens as early as step 10, indicating that the regression oracle requires very little data to start making a difference. We also observe that the more clustered rewards of the pattern step simulator allow the strategies to converge in performance to a greater degree than the higher variance rewards of the stationary step simulator. 

Most important, these experiments strongly support our expectations regarding the regression oracle's advantage when working with data containing temporal patterns.


\subsection{Forced Exploration Results}\label{sec:results-forced}

These same experiments were repeated with a forced exploration period of four pulls per arm, the results of which are presented in the right-hand side of Tables~\ref{tab:results-distribution} and~\ref{tab:results-pattern} for the stationary and pattern step simulators respectively. The lower graphs in Figures~\ref{fig:gammaonly-combined} and~\ref{fig:realsteps-combined} show the average rewards observed by the MAB strategies over time.

Results show that the relative order in performance among the six MAB strategies does not change (with the exception of $\epsilon$-decreasing regression slightly outperforming $\epsilon$-greedy regression). The use of forced exploration endures a cost of lower rewards during the forced steps with the expectation of higher rewards afterward to reclaim that cost. Looking at the Last 7 Days columns in Tables~\ref{tab:results-distribution} and~\ref{tab:results-pattern}, comparisons can be made for any of the MAB strategies regarding performance with and without forced exploration.  As it can be seen, 
forced exploration strategies generally obtain a higher reward in the last time steps than strategies without forced exploration. However, because of the lower reward early on, the average reward across the whole experiment is generally lower or even. Therefore, we conclude that if the target horizon in a given application domain is very short (even shorter than our horizon of 70), forced exploration may not provide advantage in our domain, but it may still be advantageous in scenarios with (short) horizons longer than 70.




\subsection{Results of UCBT Exploration}\label{sec:results-ucbt}

\begin{figure}[t!]
	\includegraphics[width=\columnwidth]{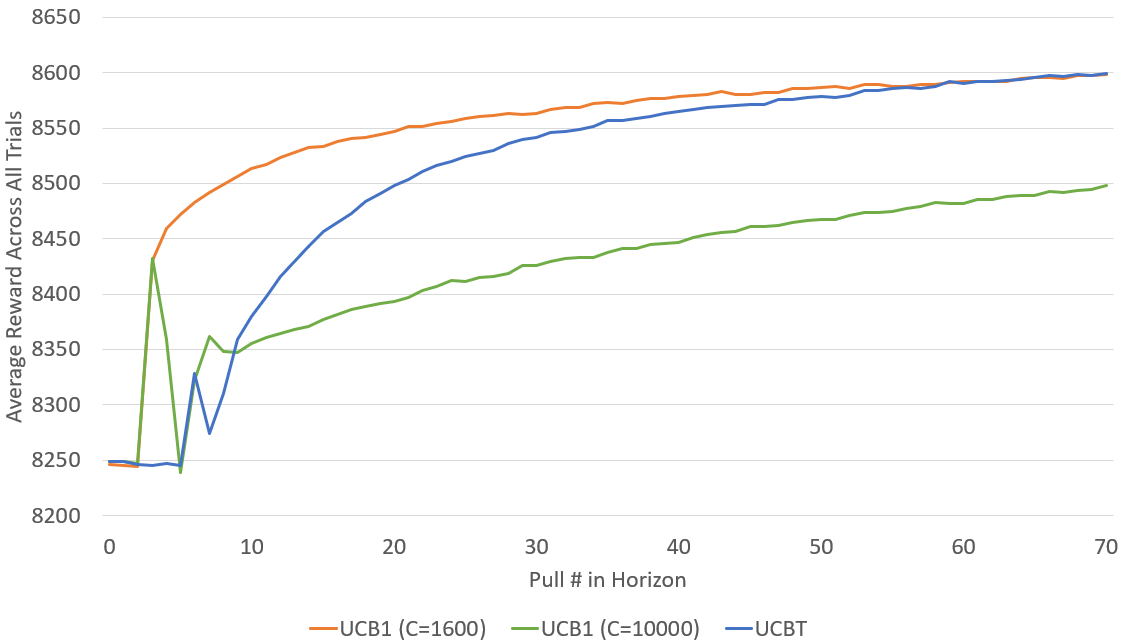}
	\centering
	\caption{Comparison among three UCB variants in the stationary step simulator. The UCBT strategy (which requires no parameter tuning) approaches the performance of the UCB1 strategy with a well-tuned C parameter and outperforms the UCB1 strategy with a poorly tuned C parameter.}
	\label{fig:ucb-comparison}
\end{figure}

In the results presented in the previous sections, it may appear that our new strategy UCBT is outperformed by UCB1. However, those results correspond to UCB1 using $C = 1600$, which was the parameter value that performed the best in our pre-experiments. However, if $C$ is not correctly tuned, UCB1 can significantly underperform. Figure~\ref{fig:ucb-comparison} displays the results of UCBT versus UCB1 with different values for $C$. We show results both for a well tuned value ($C=1600$) as well as a poorly tuned value ($C=10000$).  UCBT is scenario-agnostic and does not require any parameters or preparatory tuning. As the results demonstrate, UCB1 with an appropriate $C$ parameter excels in the early pulls and throughout the experiment, while the UCB1 strategy with the inappropriate $C$ parameter struggles. The UCBT strategy has a slower start due to the two-period forced exploration required to construct variance metrics for each arm; however, UCBT quickly catches up to match the performance of the optimal UCB1 in a relatively short horizon test. Thus, we believe UCBT is an promising strategy for real scenarios where it is not possible to tune the $C$ parameter beforehand. An example might be when we deploy our MAB strategies with real human players and find that some exhibit step distributions significantly different from those used to generate the simulators in this paper, where each would require different values for $C$.

\section{Discussion}\label{sec:discussion}




One of the main contributions of this paper is the idea of using a {\em regression oracle} in multi-armed bandit strategies. The key idea is that rather than calculating the expectation over all the previous pulls of an arm to estimate the expected reward for that arm, we build a linear regression model that takes into account all the data obtained from the problem (including pulls from other arms). We argue that in real-world bandits, where each arm might represent a possible action in a given game or a potential intervention for a human subject, often the different arms are correlated: the subject for which we are choosing an action may present general characteristics that consistently influence rewards independent of the selected arm. Therefore, by using all the available information, we are able to learn these general trends and converge to an accurate estimation of the expected reward of an arm faster than when considering only the rewards of each arm independently. Of course, linear regression is just one among the many possible regression techniques that could be used.

As part of our future work, we would like to analyze regression oracles in different stochastic bandit problem scenarios to better understand the potential benefits. 
For example, in the case where there are no temporal trends in the reward function but arm reward functions are still correlated, we hypothesize that a regression oracle will still provide the benefit of estimating the expected reward of an arm faster than traditional methods. If this is true, we should be able to replicate its effect with a simpler oracle that considers the deviation of new rewards from the mean of all rewards so far observed.
This is a hypothesis we would like to test in the future, as it could result in new bandit strategies that perform better than existing ones in short horizon situations. 

Notice that in the long horizon setting, we should see no benefits from a regression oracle unless there are temporal trends in the reward function. In the limit, the bandit strategy's reward estimates should converge to the real rewards, regardless of whether it uses data from arms together or independently. Therefore, our strategies should have the same theoretical properties as existing strategies in the limit (e.g., linear cumulative regret for $\epsilon$-greedy or logarithmic for $\epsilon$-decreasing). We would therefore like to emphasize again that the main contribution of our work is in improving the behavior of bandit strategies in very short horizon scenarios, which have been understudied in the literature, but are more realistic when applying bandits to problems involving humans.

Concerning our new strategy UCBT, this paper only showed empirical results without any theoretical bound on regret growth. Our intuition is that UCBT should still have logarithmic cumulative regret growth like UCB1, but building a proof is still part of our ongoing work. As our experiments demonstrate, however, UCBT showed superior performance than a poorly tuned UCB1. When compared to a well tuned UCB1 strategy, UCBT struggles initially, though it appears to quickly catch up in our short horizon scenario. This is interesting for two reasons. First, it means that UCBT might be more appropriate when we need to deploy an MAB strategy to a real-world problem for which we do not have sample data beforehand and cannot tune $C$ in advance. Second, UCBT might be useful beyond just standard bandit problems. For example, when algorithms such as MCTS~\cite{browne2012survey} are applied to games where rewards are not clearly bounded, the challenge of finding a $C$ that works well for all nodes in the MCTS tree might invite UCBT as an interesting alternative. We plan to study these possibilities and thoroughly analyze UCBT's theoretical properties in our future work.

Finally, we would like to reiterate that although the scenario where a bandit interacts repeatedly with the same human is often modeled as an MAB problem (as we do in this paper), it violates some of the basic assumptions of the MAB problem formulation. In particular, when an arm represents an intervention, it will by design influence the subject and thereby alter the reward function for future iterations. Strictly speaking, this is closer to reinforcement learning (as there is state) than bandits (which are stateless). However, as we have shown, it is still possible in practice to use MAB strategies in these problems with good performance.

\section{Conclusion}\label{sec:conclusion}

This paper focused on the problem of short-horizon multi-armed bandits, where we can only expect to have a small number of iterations to interact with the environment. These short-horizon problems are common in real-world situations, such as when using bandits to interact with human players in games (for example, for player modeling~\cite{gray2020player}), but they have been understudied in the literature. We presented three key ideas: regression oracles, a comparison of different exploration strategies with forced exploration in the short horizon setting, and a new variant of the UCB1 strategy called UCBT.

Our results show that regression oracles do not only help in the case where there are temporal patterns in the reward function, but also in the absence of such patterns when reward functions of different arms are still related. Our UCBT strategy was found to approach the performance of UCB1, but without any parameter that required tuning ahead of time. Regression oracles applied to standard $\epsilon$-greedy or $\epsilon$-decreasing strategies were found to be the best strategies among those we experimented with in the short horizon setting.

As part of our future work, we would like to extend the idea of regression bandits to apply them to UCB-style strategies, where the confidence term would be replaced by an estimation of the confidence bound of the linear regression estimator. We would also like to analyze the theoretical properties of both UCBT and the inclusion of a regression oracle in $\epsilon$-greedy strategies. Finally, in our current work, we have been deploying these ideas for player modeling in the context of exergames~\cite{gray2020player}.



\bibliographystyle{IEEEtran}


\end{document}